\documentclass[10pt, a4paper]{article}
\usepackage[final]{lrec2026} 

\usepackage{graphicx}
\usepackage{url}
\usepackage{alphabeta}
\usepackage{latexsym}
\usepackage{booktabs}

\title{GRDD+: An Extended Greek Dialectal Dataset with Cross-Architecture Fine-tuning Evaluation}

\name{%
\fontsize{12}{14}\selectfont\textbf{Stergios Chatzikyriakidis$^{1\dag\footnotemark[1]}$, Dimitris Papadakis$^{1\dag\footnotemark{}}$}\\
\fontsize{12}{14}\selectfont\textbf{Sevasti-Ioanna Papaioannou$^{2\dag\footnotemark{}}$, Erofili Psaltaki$^{3\dag\footnotemark{}}$}%
}

\address{
$^{1}$University of Crete,
$^{2}$University of Athens, 
$^{3}$University of Turku \\
\{stergios.chatzikyriakidis, philp0961\}@uoc.gr, lt12200018@di.uoa.gr, erofili.psaltaki@utu.fi
}

\date{}

\abstract{
We present an extended Greek Dialectal Dataset (GRDD+) that complements the existing GRDD dataset with more data from Cretan, Cypriot, Pontic and Northern Greek, while we add six new varieties: Greco-Corsican, Griko (Southern Italian Greek), Maniot, Heptanesian, Tsakonian, and Katharevusa Greek. The result is a dataset with total size 6,374,939 words and 10 varieties. This is the first dataset with such variation and size to date. We conduct a number of fine-tuning experiments to see the effect of good quality dialectal data on a number of LLMs. We fine-tune three model architectures (Llama-3-8B, Llama-3.1-8B, Krikri-8B) and compare the results to frontier models (Claude-3.7-Sonnet, Gemini-2.5, ChatGPT-5).
\\ \newline
 }

\begin{document}
\maketitleabstract

\renewcommand{\thefootnote}{\fnsymbol{footnote}}
\footnotetext[2]{These authors contributed equally to this work.}

\renewcommand{\thefootnote}{\arabic{footnote}}
\setcounter{footnote}{1}

\footnotetext{The full code for fine-tuning and the dataset \textsc{GRDD+} are available at the following anonymous link: \url{https://drive.google.com/drive/folders/1Xwfz08S8-9ZqMGd6EaNSje33LIaSE2E5?copy}.}

\section{Introduction}
Modern Greek exhibits rich dialectal variation across different geographical regions. Despite this diversity, computational resources for these dialects remain limited, constraining the study and processing of regional linguistic varieties. Meanwhile, in the current rapid advancement of Natural Language Processing (NLP), Large Language Models (LLMs) have emerged at the forefront of research and development. However, LLMs often struggle with dialectal variations in languages with less resources, e.g., in Parts of Speech (POS) tagging and dialect identification \cite{faisal2025testing}. While their dialect performance can surpass zero-shot transfer, it still falls behind the fine-tuned results \cite{faisal2024dialectbench}. These limitations significantly impact their ability to generate contextually appropriate responses across regional dialects.

This paper introduces an extended dataset GRDD+ and fine-tuning experiments across multiple model architectures. Our study aims to evaluate how model adaptation can improve dialectal performance in Greek and provide new benchmarks for dialectal NLP. The remainder of the paper is organized as follows. Section \ref{section2} reviews related work. Section \ref{sectiongrdd} describes the dataset and methodology, while Section \ref{sectionfinetun} presents the fine-tuning experiments. Section \ref{sectionresuldiscuss} reports and discusses the results. Section \ref{sectionfw} outlines future work, while Sections \ref{sectionlimit} and \ref{sectionconclus} discuss the limitations and present the conclusion and closing remarks, respectively.

\section{Related Work}
\label{section2}

Against this backdrop, resources for Modern Greek dialects remain scarce. Existing datasets include, among others, a small corpus for Griko \cite{anastasopoulos2018part}, the Cypriot Greek version of the Multi-CAST corpus of annotated spoken texts \cite{hadjidas2015multi}, and a database comprising 505 hours of recorded dialectal speech with linguistic and meta-linguistic annotations \cite{karasimos2008greed}. To our knowledge, the GRDD corpus \cite{chatzikyriakidis2023grdd} constitutes a first comprehensive effort to develop large-scale publicly available resources for Modern Greek dialects.

In parallel, several studies have emerged in the field of Greek computational dialectology, such as the development of treebanks and parsers for Eastern Cretan in the framework of Universal Dependencies \cite{vakirtzian2025dialectal}, the detection of Italian and Turkish loanwords in Greek dialects \cite{scherrerpsaltakichatzikyriakidis2025laonwords} and computational analyses of the linguistic varieties of Cappadocian, Pharasiot, and Silliot \cite{bompolas2023computational}. However, to the best of our knowledge, none of these efforts have attempted fine-tuning Large Language Models (LLMs) on Greek Dialectal Data.

\section{GRDD+ Dataset}
\label{sectiongrdd}
\subsection{Collection Methodology}

We focused on freely available dialectal data collected from the web. These include texts from blogs, websites, and publicly accessible literary sources such as songs, poems, folktales, dialogues and translations of works into the dialect by native speakers. Additionally, we collected dialectal data for certain varieties (Greco-Corsican, Griko, Heptanesian, Maniot and Pontic) from publicly available books using Optical Character Recognition (OCR) via Google Cloud Vision OCR\footnote{\url{https://cloud.google.com/vision}}, subsequently removing all book metadata and retaining only the clean dialectal text. After data collection, we performed basic preprocessing on the data, including the removal of numbers, URLs, special characters, duplicate lines and extra white spaces. 

Building upon the GRDD dataset \cite{chatzikyriakidis2023grdd}, which consists of four dialects of Modern Greek, specifically Cretan, Pontic, Northern Greek, and Cypriot Greek, the present work seeks to extend and enhance the resource. Specifically, we enrich the existing dialectal corpora and incorporate six additional Greek dialectal varieties, as detailed below.

\subsubsection{Greco-Corsican}
In the 1670s, Greek migrants from Mani settled in Cargèse, Corsica, forming a Greek-speaking community \cite{nicholas2005history}. From the 1670s to the 1960s, a span of nearly three centuries, Greek was spoken in Cargèse, in relative isolation from other Greek-speaking communities. The variety, known as Greco-Corsican, has been the subject of detailed linguistic study \cite{phardys1888yleiskarifima,blanken1951grecs,parlangeli1952dom,rexine1966vayacacos}. However, linguistic assimilation progressed rapidly, and by the 1930s only about 20 speakers of Greek remained. The language ultimately became extinct with the death of its last native speaker, Justine Voglimacci, in 1976.

\subsubsection{Griko (Southern Italian Greek)}
Griko is a Greek dialect spoken in Grecìa Salentina, southern Italy, and recognized as a minority language. Officially, Grecìa Salentina consists of 12 villages: Calimera, Carpignano Salentino, Castrignano dei Greci, Corigliano d'Otranto, Cutrofiano, Martano, Martignano, Melpignano, Sogliano Cavour, Soleto, Sternatia and Zollino. Griko together with Grecanico of Calabria, form the endangered Italiot Greek group \cite{salminen1999unesco}. Written in the Latin alphabet and only partly intelligible with Modern Greek, Griko now has fewer than 20,000 mostly elderly speakers \cite{chatzikyriakidis2010clitics}.

\subsubsection{Heptanesian}
Heptanesian is a Modern Greek dialect spoken in the Ionian Islands, including Corfu, Cephalonia, Lefkada, Zante, Ithaca, Kithira, Paxi and smaller islands such as Othoni, Antipaxi, and Antikithira \cite{kontosopoulos2000nikolaou}. These islands were under Venetian rule from the late 14th to the late 18th century. Heptanesian exhibits Venetian and Italian influences primarily in its vocabulary, phonology (e.g. intonation), and morphology (e.g. the noun suffix –aδa < Ven –ADA), with syntax largely unaffected \cite{ralli2012verbal}. Today, Heptanesian is gradually being abandoned in favor of Standard Modern Greek (SMG).

\subsubsection{Tsakonian}

Tsakonian, a highly divergent modern form of Greek, still spoken in the eastern Peloponnese, is often considered distinct enough to be classified as a separate language from the rest of Modern Greek. As the only Modern Greek dialect not descended from the Hellenistic Koine, Tsakonian represents the main exception among Modern Greek varieties, deriving more or less directly from the ancient Doric dialect \cite{joseph1987modern, mackridge2010modern}. \citeauthor{horrocks2014greek} refers to Tsakonian as a case of extreme dialectal resilience, exempt from the fundamental sound changes that shaped Modern Greek, such as the reversal of /u/ > /i/, while the dialect also exhibits numerous features that are unusual or unique compared to other Modern Greek varieties \cite{liosis2016tsakonian}.

\subsubsection{Maniot}

Maniot refers to the dialect spoken in the region of Laconian Mani. According to the traditional classification proposed by \citeauthor{hatzidakis1892einleitung}, which divides Modern Greek dialects into northern and southern groups, Maniot is categorized among the southern varieties. \citeauthor{kontosopoulos2008dialects} notes that Maniot constitutes a dialect distinct from the rest of the Peloponnesian varieties. The same view appears to be supported by \citeauthor{trudgill2003modern}, who emphasizes the distinctiveness of the Maniot dialect.

The linguistic systems that appear to share similarities with Maniot include SMG, Cretan \cite{trudgill2003modern}, Megarian, and, of course, several other Peloponnesian dialects \cite{pantelidis2001peloponnesiakos}. \citeauthor{pantelidis2001peloponnesiakos} has argued that the similarities observed between SMG and the Peloponnesian dialects result from the influence of SMG on these dialects, and not vice versa, as had previously been claimed by \citealp[among others]{mackridge1994neogreek, browning1969medieval, kontosopoulos2008dialects}.

\subsubsection{Katharevusa Greek}

Katharevusa, described as a `\,\textit{purist}\,' (literally the purifying language) variety of SMG, served as the official written language of Greece from the establishment of the modern Greek nation-state until 1976 \cite{joseph1987modern, mackridge2010modern}. This language variety was the intermediate solution during the language controversy\footnote{The language controversy, which originated in the 1760s, re-emerged with the establishment of the modern Greek state (1830) through the debate over which variety should serve as the official language of the newly independent nation \cite{mackridge2010modern}. Katharevousa emerged as a kind of compromise between adopting Ancient Greek and the spoken form of SMG as the national language \cite{horrocks2014greek}.}, and it was mostly used in written texts \cite{mackridge2010modern} (for a contrasting view that argues that conditions of diglossia were developed between Katharevousa and SMG here: \citealp{joseph1987modern}). Katharevusa combined elements of both Ancient and Modern Greek, retaining much of the classical vocabulary and morphology while introducing intermediate forms such as εἴμεθα `\,\textit{we are}\,'  and ἦτον `\,\textit{he/she/it was}\,'. Syntactically, it was closer to SMG, using constructions like νὰ + finite verb and the negative δὲν, yet it preserved many ancient participial structures absent from the spoken language \cite{mackridge2010modern, horrocks2014greek}.

\subsubsection{CretDeiAdv (Cretan deictic adverbs)}
During \citeauthor{psaltaki2025deixis}’s master’s thesis, she studied Cretan adverbs expressing deixis. At the time, no dialectal corpus was available, so she created a corpus containing examples of adverbs denoting here and there. The corpus combines texts from the Cretan Renaissance (15th–17th c.) collected by Kaklamanis (2020) and 62 folklore books (1876–2020). The resulting resource, CretDeiAdv \cite{psaltaki2025deixis}, is ideal for researchers interested in Cretan adverbs, particularly deictic expressions, and is being offered to the research community for further study.

\subsection{Dataset Statistics and Characteristics}
The GRDD original corpus comprises four main Greek dialects: Pontic Greek, Cretan Greek, Cypriot Greek, and Northern Greek. We used the term \textbf{Pontic Greek} to refer to the dialect spoken today in modern Greece, although a form of Pontic, Romeyka Pontic, is still spoken in some villages of Trabzon and surrounding areas in modern Turkey \cite{sitaridou2012cultural}. \textbf{Cretan Greek} is spoken on the island of Crete and is derived from Koine Greek \cite{mackridge1985moderngreek}. \textbf{Cypriot Greek} is spoken primarily by Greek Cypriots, as well as by some Turkish Cypriots. The previous version of the corpus on \textbf{Northern dialects} included data only from Kozani and Grevena, but we have now extended it to also include Lesbos, Samothrake and Thrace reflecting the broader scope of Northern dialects. This is something worth mentioning even though we will not discuss the original corpus dialects \cite{chatzikyriakidis2023grdd} in detail. 

With the dialects of the GRDD original corpus that are shown in table \ref{tab:corpus-wordcounts}, the addition of new data results in substantial growth across several dialects. \textbf{Pontic Greek} increases moderately, with the new words contributing roughly +8.1\% to the original 867,935, for a total of 938,220 words. \textbf{Cretan Greek} experiences a more pronounced expansion, adding 583,808 words, an increase of 64.8\%, bringing its total to 1,484,203 words. \textbf{Cypriot Greek} grows modestly, with the new words contributing roughly 2.1\% to the original 1,345,849, for a total of 1,374,024 words. \textbf{Northern Greek}, initially the smallest of these dialects, more than triples in size with the new additions, rising by about 260.1\% to reach 119,894 words. This growth improves dataset coverage, supporting more robust cross-dialectal analyses and computational modeling. 

The newly added data includes several varieties that were not present in the original dataset. \textbf{Katharevousa} dominates with 1,515,982 words. \textbf{Tsakonian} also has a substantial representation with 442,512 words, highlighting the effort to document this highly endangered dialect. \textbf{Grico}, an Italo-Greek minority language, contributes 366,889 words, providing important coverage for a Greek variety outside Greece.

Smaller dialects include \textbf{Heptanesian} (50,311 words), \textbf{Maniot} (30,692 words) and \textbf{Greco-Corsican} (5,026 words). There is also \textbf{CretDeiAdv} (47,186 words), which represents a specialized subcorpus focusing on deixis adverbs in Cretan Greek. Despite their smaller size, these additions are valuable for the preservation and analysis of minority or regionally restricted varieties and for studies of specific linguistic phenomena. These additions improve the coverage of minority, regional and specialized varieties, supporting more robust cross-dialectal analyzes and computational modeling.The overall size of the corpus has approximately doubled following the incorporation of the new data. Table~\ref{tab:corpus-wordcounts} shows the distribution.

\begin{table*}[t]
\centering
\small
\begin{tabular}{@{}lrrr@{}}
\toprule
\textbf{Dialect/Variety} & \textbf{GRDD Word count} & \textbf{New Word count} & \textbf{GRDD+ Word Count} \\
\midrule
Pontic         & 867,935   & 70,285   & 938,220   \\
Cretan         & 900,395   & 583,808  & 1,484,203 \\
Cypriot        & 1,345,849 & 28,175   & 1,374,024 \\
Northern       & 33,292    & 86,602   & 119,894   \\
Katharevousa   & --        & 1,515,982 & 1,515,982 \\
Tsakonian      & --        & 442,512  & 442,512   \\
Grico          & --        & 366,889  & 366,889   \\
Heptanesian    & --        & 50,311   & 50,311    \\
Maniot         & --        & 30,692   & 30,692    \\
Greco-Corsican & --        & 5,026    & 5,026     \\
CretDeiAdv     & --        & 47,186   & 47,186    \\
\midrule
\textbf{Total} & 3,147,471 & 3,227,468 & 6,374,939 \\
\bottomrule
\end{tabular}
\caption{Word counts in the original GRDD corpus, newly added words for existing and new dialects/varieties, and total word counts in GRDD+ per dialect/variety.}
\label{tab:corpus-wordcounts}
\end{table*}

\section{Fine-tuning Methodology}
\label{sectionfinetun}

\subsection{Fine-tuning Data Construction}

We constructed a dialectal fine-tuning dataset from raw text corpora representing four Greek regional dialects from the GRDD collection: Cretan, Pontic, Northern Greek, and Cypriot Greek. To create structured training examples from the raw text, we used a sliding window approach:

\begin{enumerate}
    \item Text is split into chunks of 100 words.
    \item Chunks with at least 50 words are turned into prompt-completion pairs:
    \begin{itemize}
        \item \textbf{Longer chunks ($\ge$ 80 words)}: Split in half, first half is the prompt, second half is the completion
        \item \textbf{Shorter chunks (50-79 words)}: The full chunk is the completion
    \end{itemize}
    \item Each example starts with a dialect instruction in Greek (e.g., "Γράψε στην κρητική διάλεκτο:" for Cretan).
    \item We randomly pick from multiple instruction templates per dialect.
\end{enumerate}

This gave us 20,116 training examples across all four dialects, saved as JSONL files. Table~\ref{tab:dataset-stats} shows the distribution.

For Cypriot Greek, we combined two separate corpora: a subset of our publicly available corpus (5,625 examples from 562,522 words) and the ΑΠΟαποικιοΠΟΙΗΣΗ corpus \citep{achilleos} (964 examples from 96,410 words), used with permission from the authors.

\begin{table*}[t]
\centering
\begin{tabular}{@{}lrrr@{}}
\toprule
\textbf{Dialect} & \textbf{Words} & \textbf{Examples} & \textbf{\%} \\
\midrule
Cretan & 900,395 & 9,004 & 44.8\% \\
Pontic & 418,997 & 4,190 & 20.8\% \\
Northern & 33,292 & 333 & 1.7\% \\
Cypriot (public) & 562,522 & 5,625 & 28.0\% \\
Cypriot (ΑΠΟαποικιοΠΟΙΗΣΗ) & 96,410 & 964 & 4.8\% \\
\midrule
\textbf{Total} & \textbf{2,011,616} & \textbf{20,116} & \textbf{100\%} \\
\bottomrule
\end{tabular}
\caption{Fine-tuning dataset distribution}
\label{tab:dataset-stats}
\end{table*}
The distribution reflects the varying availability of high-quality dialectal resources in GRDD, with Cretan (44.8\%) and the combined Cypriot data (32.8\%) being well-represented, followed by Pontic (20.8\%) and Northern Greek (1.7\%). We preserved this natural distribution to maximize the use of available dialectal data, though we acknowledge this imbalance as a potential limitation that may affect relative performance across dialects.

\subsection{Models and Training Setup}
\paragraph{Base Models}
We fine-tune three models:

\begin{itemize}
    \item \textbf{Llama-3-8B}: Meta's instruction-tuned multilingual model.
    \item \textbf{Llama-3.1-8B}: Enhanced version with extended context (128k tokens).
    \item \textbf{Krikri-8B}: Greek-specialized model built on Llama-3.1-8B, trained on 56.7B Greek tokens, the premier LLM for the Greek language \citep{krikri}. 
\end{itemize}

\paragraph{LoRA Configuration}

We use LoRA \citep{hu2021lora} for efficient fine-tuning. Table~\ref{tab:lora-params} shows our settings.

\begin{table}[h]
\centering
\begin{tabular}{@{}ll@{}}
\toprule
\textbf{Parameter} & \textbf{Value} \\
\midrule
LoRA Rank ($r$) & 16 \\
LoRA Alpha ($\alpha$) & 32 \\
LoRA Dropout & 0.1 \\
Target Modules & q\_proj, k\_proj, v\_proj, \\
                & o\_proj, gate\_proj, \\
                & up\_proj, down\_proj \\
Trainable Parameters & $\sim$0.8\% of base model \\
\bottomrule
\end{tabular}
\caption{LoRA configuration.}
\label{tab:lora-params}
\end{table}

\paragraph{Training Setup}

Table~\ref{tab:training-params} shows our training hyperparameters.

\begin{table}[h]
\centering
\begin{tabular}{@{}ll@{}}
\toprule
\textbf{Hyperparameter} & \textbf{Value} \\
\midrule
Epochs & 3 \\
Batch Size per Device & 2 \\
Gradient Accumulation Steps & 8 \\
Effective Batch Size & 16 \\
Learning Rate & 3e-4 \\
LR Scheduler & Cosine \\
Warmup Steps & 100 \\
Optimizer & AdamW \\
Weight Decay & 0.01 \\
Max Gradient Norm & 1.0 \\
Precision & bfloat16 \\
Max Sequence Length & 512 tokens \\
\bottomrule
\end{tabular}
\caption{Training hyperparameters.}
\label{tab:training-params}
\end{table}

All experiments ran on AWS ml.p4d.24xlarge instances with NVIDIA A100 GPUs (40GB). Training took 4-6 hours per model, with peak memory under 35GB per GPU.

\subsection{Evaluation}
We compare our three base models, their three fine-tuned versions, and three frontier models,  Claude-3.7-Sonnet, Gemini-2.5, and ChatGPT-5. For each dialect, we use 7 different prompts (short story, 3 medium stories, long story, dialogue, creative writing), giving 7 generations per model. Given that we have a total of nine models (3 fine-tuned + 3 base models + 3 frontier models), we have 63 generations per dialect.
Native speakers evaluated the  generated texts on a 5-point scale shown in (Table~\ref{tab:eval-scale}).

\begin{table}[h]
\centering
\small
\begin{tabular}{@{}cl@{}}
\toprule
\textbf{Score} & \textbf{Description} \\
\midrule
5 & Απόλυτα φυσικό - Native-level \\
4 & Πολύ φυσικό - Minor issues \\
3 & Μέτρια φυσικό - Noticeable problems \\
2 & Αφύσικο - Significant problems \\
1 & Εντελώς αφύσικο - Not dialectal \\
\bottomrule
\end{tabular}
\caption{Native speaker evaluation scale.}
\label{tab:eval-scale}
\end{table}

\begin{table*}[t]
\centering
\small
\begin{tabular}{@{}lcccccccc@{}}
\toprule
\textbf{Model} & \multicolumn{2}{c}{\textbf{Cretan}} & \multicolumn{2}{c}{\textbf{Cypriot}} & \multicolumn{2}{c}{\textbf{Pontic}} & \multicolumn{2}{c}{\textbf{Northern}} \\
\cmidrule(lr){2-3} \cmidrule(lr){4-5} \cmidrule(lr){6-7} \cmidrule(lr){8-9}
& \textbf{Mean} & \textbf{SD} & \textbf{Mean} & \textbf{SD} & \textbf{Mean} & \textbf{SD} & \textbf{Mean} & \textbf{SD} \\
\midrule
Llama-3-8B (base) & 1.15 & 0.49 & 1.52 & 1.03 & 1.11 & 0.52 & 1.32 & 0.89 \\
Llama-3-8B (fine-tuned) & 3.67 & 1.23 & 3.23 & 1.20 & 2.83 & 1.13 & 2.84 & 1.30 \\
\midrule
Llama-3.1-8B (base) & 1.13 & 0.45 & 1.38 & 0.78 & 1.00 & 0.00 & 1.30 & 0.92 \\
Llama-3.1-8B (fine-tuned) & 3.20 & 1.41 & \textbf{3.51} & 1.15 & \textbf{2.86} & 1.12 & 3.10 & 1.28 \\
\midrule
Krikri-8B (base) & 1.28 & 0.63 & 1.95 & 1.35 & 1.06 & 0.33 & 1.41 & 1.05 \\
Krikri-8B (fine-tuned) & 2.80 & 1.35 & 3.36 & 1.27 & 2.49 & 1.16 & 3.22 & 1.28 \\
\midrule
ChatGPT-5 & 2.49 & 1.37 & 3.36 & 1.08 & 2.14 & 0.96 & 3.54 & 0.89 \\
Claude-3.7-Sonnet & \textbf{3.79} & 1.23 & 3.48 & 1.13 & 2.83 & 1.06 & \textbf{3.86} & 1.10 \\
Gemini-2.5-Pro & 1.63 & 0.92 & 2.47 & 1.20 & 1.06 & 0.33 & 2.02 & 1.00 \\
\bottomrule
\end{tabular}
\caption{Native speaker evaluation scores (1-5 scale) across dialects and models. Mean and standard deviation (SD) reported for each dialect. Cretan had 16 raters, Cypriot 19 raters, Northern 9 raters and Pontic 5 raters.}
\label{tab:results}
\end{table*}

\section{Results and Discussion}
\label{sectionresuldiscuss}
The results of our evaluation are shown in Table \ref{tab:results}. 
Inter-rater reliability was assessed using multiple metrics and the results are presented in Table~\ref{tab:interrater}. Krippendorff's Alpha ranged from 0.37 to 0.55 across dialects, which indicates fair to moderate agreement on absolute scores. ICC(3,1), basically a two-way mixed effects model treating raters as random and items as fixed, yielded values between 0.87 and 0.96, demonstrating excellent consistency in relative rankings. Weighted Cohen's Kappa, calculated as the average across all rater pairs and accounting for ordinal distance between ratings, ranged from 0.39 to 0.54, falling between the other two metrics in sensitivity to absolute differences.

In terms of individual dialects, Cretan showed the highest agreement across all metrics (Krippendorff's $\alpha$=0.55, ICC(3,1)=0.96, weighted $\kappa$=0.54).  Cypriot, on the other hand,  showed the lowest Krippendorff's Alpha (0.37) and weighted Kappa (0.39), even though it  maintains high consistency in relative rankings (ICC(3,1)=0.95). This pattern may reflect greater dialectal variation and/or also point to the larger, and potentially more diverse, rater pool for Cypriot (19 raters versus 5-16 for other dialects). Pontic showed substantially higher exact agreement (39.7\%), with ratings also clustering at lower points in the scale compared to the other dialects. 

The high ICC(3,1) values demonstrate that despite differences in scale usage, raters consistently agreed on which texts were better or worse, a critical requirement for valid model comparisons. These results validate the use of averaged ratings while acknowledging the inherent subjectivity in dialectal quality assessment.

\begin{table*}[t]
\centering
\begin{tabular}{@{}lcccc@{}}
\toprule
\textbf{Metric} & \textbf{Northern (9 raters)} & \textbf{Cretan (16 raters)} & \textbf{Pontic (5 raters)} & \textbf{Cypriot (19 raters)} \\
\midrule
Krippendorff's $\alpha$ & 0.429 & \textbf{0.545} & 0.425 & 0.373 \\
ICC(3,1) & 0.442 & \textbf{0.551} & 0.451 & 0.384 \\
Weighted $\kappa$ (avg) & 0.449 & \textbf{0.542} & 0.435 & 0.389 \\
\midrule
Exact agreement (\%) & 8.2 & 1.6 & 39.7 & 0.0 \\
\bottomrule
\end{tabular}
\caption{Inter-rater reliability across dialects. Krippendorff's $\alpha$, ICC(2,1), and weighted $\kappa$ are appropriate for ordinal scales and indicate fair to moderate agreement (0.37--0.55). Exact agreement percentages show expected low values for subjective multi-rater evaluations, except for Pontic which has a fair exact agreement consensus.}
\label{tab:interrater}
\end{table*}
There are many interesting things to note about the results both in terms of fine-tuning and model choice. The most straightforward comparison is between the base versions of Llama and frontier models like GPT5, Claude 3.7 and Gemini 2.5-Pro. Llama base models including Krikri have close to zero dialectal knowledge while the frontier models range seem to possess dialectal knowledge to varying degrees, from Gemini to Claude.  

Another dimension in the discussion concerns the relation between Llama base and fine-tuned versions. It is clear that all fine-tuned models are much better than the base models, showing an increase of between 1.5-2 points approximately in their fine-tuned versions. 

Comparing the fine-tuned models, we notice a number of interesting things but not an across the board clear picture. The first thing that stands out is that Llama-Krikri, the only model which is explicitly trained in Modern Greek, does not show the best performance among the three. Llama-Krikri only performs better in the generation of Northern Greek, scores second for Cypriot, and third (last) in the other two dialects, i.e. Cretan and Cypriot. This might be an indication that the other two models are more flexible in learning the new varieties than Krikri, despite the latter being explicitly trained on Modern Greek. 

Finally, comparing the fine-tuned 8B with the three frontier models, a number of interesting findings also arise there. First of all, Claude 3.7 is consistently high-performing, topping the Northern and Cretan category, and being second, very close to the first, for Cypriot and Pontic. It is important to note here that the newer Claude versions (4 onwards) have lost their dialectal capabilities to some extent, and this is one of the reasons that we used this model rather than the new ones. Why this has happened and to what extent, is an issue that warrants further investigation which will not be undertaken here. GPT5 is performing decently consistent, giving quite good performances for Cypriot and Northern and rather mediocre for the other two. Gemini has consistently mediocre to poor performance ranging from 2.47 to 1.06. 

In terms of the individual dialects, we notice that the higher scores are given to Northern and Cretan respectively, followed by Cypriot and Pontic. An interesting question here concerns whether this cline has anything to do with the distance of these individual dialects to the dominant variety, Modern Greek, that all models have at least some knowledge of. Impressionistic intuitions about these dialects dictate that indeed Northern and Cretan are closer to the dominant variety, while Cypriot and lastly Pontic are farther away.\footnote{In traditional Greek dialectology, there is a distinction between idioms and dialects. Basically, idioms were varieties that were closer to the dominant but not that far away to be considered dialects, and dialects were varieties that were farther away from the dominant to be considered idioms. Northern and Cretan were usually considered idioms, Pontic and Cypriot dialects} Of course, the issue of linguistic distance is largely unexplored in Greek varieties, but it would be interesting to see whether these results here, correlate with some notion of distance between the dominant variety and the respective dialects. 

Lastly, the relationship between training data size and model performance across the four dialects is shown to be quite intriguing. Cretan is the first  in size with 9,004 examples (44.8\%) and performs well (2.80-3.79), while Cypriot has 6,589 examples (32.8\% of the dataset) and all three fine-tuned models scoring above 3.0, with one fine-tuned model (Krikri-8B at 2.80) falling  below the 3.0 threshold. Pontic has 4,190 examples (20.8\%) but consistently scores lowest (2.14-2.86), with all three fine-tuned models failing to reach 3.0. Surprisingly, Northern Greek with only 333 examples (1.7\%), manages to achieve strong scores (2.84-3.86), with only one fine-tuned model (Llama-3-8B at 2.84)  below 3.0. The fact that Cypriot is the only dialect where all fine-tuned models consistently manage to break the 3.0 might indicate the benefits of a large dataset  that is  the combination of two diverse corpora, providing better coverage. The Northern results remain notable, as they despite having the least data by far, it matches Cretan's consistency better than Pontic does, with 12 times more training examples. This pattern might be an indication of linguistic distance from Standard Modern Greek, data quality differences, or a combination of the two. 
\section{Future Work}
\label{sectionfw}
With respect to dataset creation, we would like to do a thorough evaluation of the data collected to come up with potentially more fine-grained categories. For example, Cypriot Greek data is comprised by data from a number of genres including blog posts, literature written in Cypriot,  traditional songs and riddles, as well as some limited scientific texts in Cypriot. Classifying into more specific genres will potentially help research in other fields of Linguistics. On that note, sociolinguistic considerations with respect to the current diglossic situation in Cypriot, as well as the use of a Cypriot Koine are also issues that might be benefited from some of our data, identifying particulars patterns that are the result of code-switching or other sociolinguistically relevant markers (e.g. Shibolleth markers of local varieties \citep{shibboleth}). 

One of our immediate plans of continuing this work concerns the fine-tuning on the six newly added varieties (Greco-Corsican, Griko, Heptanesian, Tsakonian, Maniatika, Katharevusa). This would provide a more comprehensive coverage across all GRDD+ dialects. We also plan to test a number of additional architectures (e.g. Mistral, Gemma) and parameter-efficient methods, as well as exploring multi-dialect models that have the ability to handle all varieties simultaneously.

We plan to develop automatic evaluation metrics for dialectal quality in order to enable research. Task-based evaluation (summarization, question-answering, translation) can complement our generation approach, while, at the same time, provide insights into dialectal comprehension capabilities.

Lastly, we plan to continue expanding the GRDD dataset by enhancing both the current dialectal corpora and adding new dialects to broaden its coverage.

\section{Limitations}
\label{sectionlimit}

\textbf{Dataset imbalance.} The dataset used for fine-tuning is very imbalanced.  Cretan comprises 44.8\% (9,004 examples), while  Northern Greek only 1.7\% (333 examples). Of course, this reflects the varying resource availability, and it is understandable to some extent,  but can affect the relative performance across dialects. 

\textbf{Evaluation subjectivity.} The results we have show   moderate agreement levels (Krippendorff's $\alpha$=0.37--0.55). This reflects an inherent subjectivity in these type of naturalness  judgments. ICC(3,1) values (0.87--0.96) indicate excellent consistency in relative rankings, but, however, the variability in ratings might suggest a  need for more structured evaluation protocols or even rater training. Lastly, our evaluation uses only 7 prompts per dialect, primarily narrative tasks, and as such do not capture a fuller  range of  usage contexts.

\textbf{Limited scope.} We fine-tuned  three 8B models with a single LoRA configuration and then compared them against three frontier models. Furthermore, the unexpected underperformance of Krikri-8B, despite its Greek-specific training, merits further investigation.

\textbf{Linguistic distance.} Our discussion of performance relative to linguistic distance from Standard Modern Greek remains qualitative. Greek dialectology lacks standardized distance metrics that would enable systematic testing of this relationship.

\textbf{Sociolinguistic factors.} We do not account for diglossia, code-switching, register, genre, or within-dialect variation (e.g., local varieties vs. Cypriot Koine \citep{tsiplakou}).

\section{Conclusion}
\label{sectionconclus}

We presented GRDD+, an extended Greek dialectal dataset that includes data from 10 varieties of Greek. 4 of the dialects were part of the existing GRDD dataset (Cretan, Cypriot, Pontic, Northern) and have been expanded in terms of coverage, and  six new varieties (Greco-Corsican, Griko, Heptanesian, Tsakonian, Maniot, and Katharevusa) were added. This is a good basis for a  comprehensive resource for Greek dialectal NLP.

We then experimented with a number of fine-tuning experiments using  three 8B parameter models (Llama-3, Llama-3.1, and Krikri). The results show that targeted dialectal fine-tuning improves generation quality, showing gains of 1.5-2 points on a 5 point scale of dialectal naturalness. This is even true with relatively modest amounts of training data, as evidenced by Northern Greek, that achieves strong performance (2.84-3.86) despite having only 333 training examples.

Comparison with frontier models shows a rather nuanced performance picture. Claude-3.7-Sonnet achieves the highest scores for Cretan (3.79) and Northern Greek (3.86), while the fine-tuned Llama-3.1-8B outperforms all models on Cypriot (3.51) and Pontic (2.86). This is an indication that specialized fine-tuning can enable smaller models to exceed frontier model performance on specific dialects. ChatGPT-5 shows solid, albeit inconsistent performance across dialects, and Gemini-2.5-Pro is  underperforming across all dialects. Notably, base Llama models (including the Greek-specialized Krikri) show near-zero dialectal capabilities, highlighting the critical importance of dialectal training data.

A number of other findings warrant further investigation: (1) Krikri-8B fine-tuned is underperforming  relative to multilingual Llama models despite its Greek-specific training, (2) there is a  non-linear relationship between training data size and performance (Northern outperforming Pontic despite having 12 times less data), and (3) the correlation between dialectal performance and linguistic distance from Standard Modern Greek.

We believe that this dataset and our findings can function  as a solid foundation for future work on Greek dialectal NLP, since we show  that even small amounts of high-quality dialectal data can enable effective fine-tuning. We really hope that this resource will enable research not only in NLP but also in sociolinguistics, dialectology, and language documentation for Greek and other languages with rich dialectal variation.
\section{Acknowledgments}
We thank Andri Achilleos, Spyros Armostis, and Elena Sokratous for granting permission to use data from the ΑΠΟαποικιοΠΟΙΗΣΗ corpus. We also thank Panos Marneris for giving us permission to scrape and use the Tsakonian data found in his website.  Erofili Psaltaki received funding from the European Union’s Horizon Europe research and innovation program under the Marie Skłodowska-Curie grant agreement No 101177564—HAIF. Co-funded by the European Union. Views and opinions expressed are however those of the author(s) only and do not necessarily reflect those of the European Union or the European Research Executive Agency (REA). Neither the European Union nor the granting authority can be held responsible for them. Stergios Chatzikyriakidis  gratefully acknowledges funding from Amazon (project: Neural-Symbolic Integration
for Enhanced Natural Language Processing (NIELS)) that provided computational support for the fine-tuning experiments described in the paper. Stergios Chatzikyriakidis is also partially funded  by the European Union (ERC ADG, PhylProGramm, 101096554). Views and opinions expressed are however those of the author(s) only and do not necessarily reflect those of the European Union or the European Research Council Executive Agency. Neither the European Union nor the granting authority can be held responsible for them. 

\section{Bibliographical References}\label{sec:reference}
\bibliographystyle{lrec}
\bibliography{lrec2026}

\end{document}